\documentclass{article}
\usepackage{ijcai19}

\usepackage{times}
\usepackage{soul}
\usepackage{url}
\usepackage[hidelinks]{hyperref}
\usepackage[utf8]{inputenc}
\usepackage[small]{caption}
\usepackage{graphicx}
\usepackage{amsmath}
\usepackage{booktabs}
\usepackage{mathrsfs}
\usepackage{times}
\usepackage{fancyhdr,graphicx,amsmath,amssymb}
\usepackage[ruled,vlined]{algorithm2e}
\usepackage{multirow}
\usepackage{diagbox}

\title{Incremental Few-Shot Learning for Pedestrian Attribute Recognition}

\author{Liuyu Xiang$^1$ \and Xiaoming Jin$^{1*}$ \and Guiguang Ding$^1$\footnote{ Corresponding authors}\and Jungong Han$^2$\And Leida Li$^3$ \\ 
\affiliations
$^1$School of Software, Tsinghua University, Beijing, China\\
$^2$WMG Data Science, University of Warwick, CV4 7AL Coventry, United
Kingdom \\
$^3$ School of Artificial Intelligence, Xidian University, Xi'an 710071, China\\
\emails
xiangly17@mails.tsinghua.edu.cn, \{xmjin,dinggg\}@tsinghua.edu.cn \\
jungonghan77@gmail.com, reader1104@hotmail.com \\
}

\begin{document}

\maketitle

\begin{abstract}
Pedestrian attribute recognition has received increasing attention due to its important role in video surveillance applications. However, most existing methods are designed for a fixed set of attributes. They are unable to handle the incremental few-shot learning scenario, i.e. adapting a well-trained model to newly added attributes with scarce data, which commonly exists in the real world. In this work, we present a meta learning based method to address this issue. The core of our framework is a meta architecture capable of disentangling multiple attribute information and generalizing rapidly to new coming attributes. 
By conducting extensive experiments on the benchmark dataset PETA and RAP under the incremental few-shot setting, we show that our method is able to perform the task with competitive performances and low resource requirements.



\end{abstract}

\section{Introduction}
Pedestrian attribute recognition, aiming at recognizing human part attributes such as age, gender and clothing, is in growing demand for its wide applications in visual surveillance such as person re-identification, social behavioral analysis, and face verification. In the deep learning era, various methods based on either CNN or RNN have been proposed, and have achieved great success in this area. However, in real-world visual surveillance applications, the demand is usually dynamic and evolving, that is, the set of human part attributes that need to be recognized can be changed or enlarged over time. 

\begin{figure}[!h]
\setlength{\belowcaptionskip}{-5pt}
\centering
\includegraphics[width=0.48\textwidth]{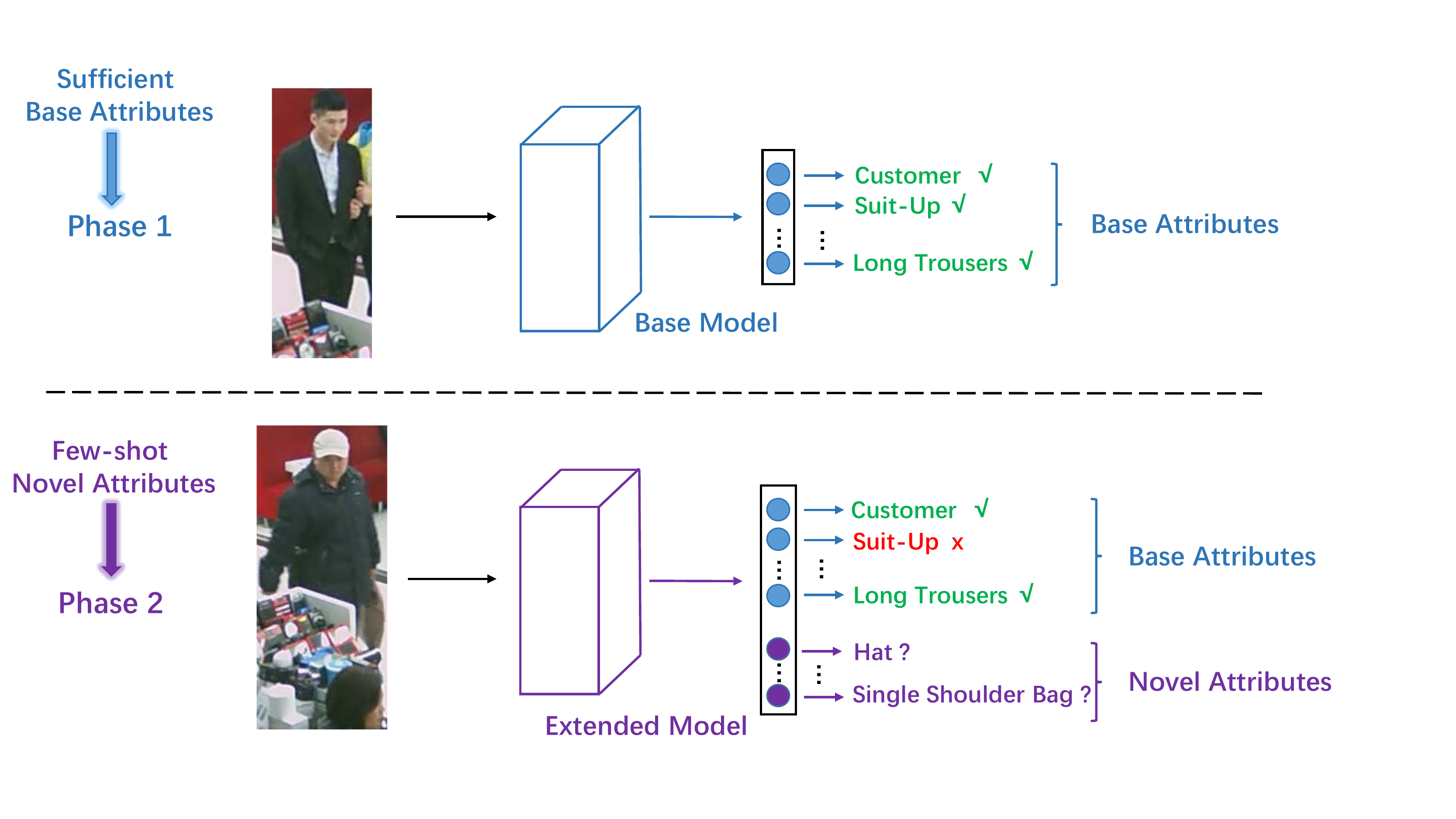}
\caption{Illustration of Incremental Few-shot Learning for Pedestrian Attribute Recognition}
\label{illustration}
\end{figure}

Consider the example in Figure \ref{illustration}. During phase one, a model is trained (which we call base model) to recognize the existence of several attributes such as \emph{'Customer'}, \emph{'Suit-up'}, and \emph{'Long Trousers'}. Then during phase two, a few-shot dataset annotated with some new attributes is provided. Our goal is to develop an efficient algorithm that can utilize both the few-shot data and the already well-trained base model, and produce an extended model that can not only successfully recognize the new attributes such as \emph{'hat'} and \emph{'Single Shoulder Bag'}, but also retain the capability of classifying old attributes such as \emph{'Customer'} and \emph{'Suit-up'}.
For clarity we denote the previous well-recognized attributes as \textbf{base attributes}, and the new coming attributes as \textbf{novel attributes}. 


Given the scenarios described above, two straight-forward solutions would be finetuning and re-training. However, finetuning, i.e. adding randomly initialized output layers to the previous model and tuning the whole network on the few-shot examples, is prone to overfitting on novel attributes, and may even lead to catastrophic forgetting on base attributes\cite{goodfellow2013empirical}. 
Re-training, i.e. discarding the previous well-trained model and training from scratch using both base and novel data, not only requires to store the previous large-scale training data, but is also time-consuming. Thus it is infeasible to re-train in most applications where the time and memory are limited.
Thus, we need to find a solution, that given the previously well-trained model and without storing the previous training data, can efficiently extend the previous model so that it is able to recognize both the base and novel attributes. 
It is worth to note that this task is challenging, since it involves incremental learning, few-shot learning and multi-label learning. Simply applying or combining any technique from only one of these areas is insufficient to solve this complex problem. 

In order to address the challenges described above, we propose to extend the base model with an extra module called Attribute Prototype Generator Module (APGM). The APGM can be regarded as a higher-level meta-learner that extracts the multiple-attribute information from the feature embedding, and produces discriminative Attribute Prototype Embedding (APE). Then the classification weights for the novel attributes are acquired based on APE. 
Further, if the \emph{base attributes} of the few-shot dataset are also annotated, we put forward another module called Attribute Relationship Module (ARM) that transfers knowledge from the base classification weights to the novel weights based on attributes co-occurrence, to further improve the performances of the generated novel classification weights.

To verify that our proposed method is capable of performing incremental few-shot learning for pedestrian attribute recognition, we conduct experiments on two benchmark datasets for pedestrian attribute recognition: PETA and RAP. We first transform these datasets into incremental few-shot case, and show by experiments that our method performs superior to the baselines with less time and memory consumption.

Besides, we also discuss the internal relationships between our method and other two contemporary meta learning methods. We explain intuitively the reason why our method is more suitable for the multi-label few-shot case, and demonstrate experimentally that simply extending the multiclass state-of-the-art few-shot meta learning algorithms into their multi-label counterparts will result in poor performances under this context. 


In summary, we
make the following contributions in this paper:
\begin{itemize}
    \item We raise a practical issue termed as incremental few-shot  (multi-label) attribute learning which commonly exists in real-word applications.
    \item We put forward a meta learning based method that achieves competitive performances. 
    \item Our method demonstrates the capability of fast adaptation with low annotation and memory requirements which is desirable in low-resource on-device scenarios.
\end{itemize}


\section{Related Work}

\subsection{Pedestrian Attribute Recognition}
Pedestrian attributes have been used as soft biometric traits, and their recognition has been intensively studied due to its wide applications in video surveillance. Traditional methods usually involve extracting hand-crafted features and optimizing classifiers such as Support Vector Machine (SVM) \cite{prosser2010person}. However, due to the limited representation capability of these hand-designed features, these traditional methods' performances still need to be improved.

With the development of automatic representation learning, deep architectures based on CNN, CNN-RNN and ConvLSTM have been proposed. In particular, \cite{sudowe2015person} first adopts CNN that is trained end-to-end with independent layers for each single attribute. \cite{li2015multi} proposes DeepMAR with a weighted sigmoid cross-entropy loss to utilize the attribute distribution prior information to handle the class-imbalance problem. 
Later, \cite{wang2017attribute} proposes a CNN-RNN based encoder-decoder framework, and \cite{zhao2018grouping} further improves recurrent learning via attribute grouping. Apart from designing novel network architectures, \cite{li2017learning} exploits a  Spatial Transformer Networks (STN) to extract multi-scale context-aware discriminative features, where \cite{zhu2017multi} divides the human body into several parts for more accurate attribute recognition. Efforts have also been made to model the local or hierarchical relationships where techniques such as attention mechanism and automatic gating \cite{liu2017hydraplus,jin2018automatic} are exploited.

\subsection{Few-Shot Learning and Meta Learning}
Meta learning, also known as \emph{learning to learn} \cite{thrun2012learning}, has achieved remarkable success in few-shot learning and other applications \cite{ravi2016optimization,finn2017model,snell2017prototypical,xiang2019adaptive}. 

The key idea of meta learning is to learn a higher-level parameterized network (which we call meta-learner), to guide an ordinary network (which we call learner) to learn. This is often achieved by the so-called episodic training, which enables the meta-network to generalize to few examples by consistently mimicking the few-shot scenarios. Among these methods, \cite{finn2017model} learns an initialization of a neural network capable of fast adaptation, where \cite{ravi2016optimization} utilizes an LSTM as meta-learner to guide the learner to update parameters using the gradient information. \cite{snell2017prototypical} and \cite{sung2018learning} approach few-shot learning via metric learning, where the former learns an embedding function followed by KNN, and the latter learns the metric function as well. Dynamical weight methods including \cite{bertinetto2016learning,qi2018low,gidaris2018dynamic} propose to train a meta-network to produce parameters of a base learner, so that the produced parameters (fast weights) can quickly adapt to few examples.

However, most existing meta learning methods are targeted for multi-class few-shot classification, and is not applicable to our \textbf{multi-label} setting where complex attribute relationships and co-occurrence exist. 


\begin{figure*}[!h]
\setlength{\abovecaptionskip}{-7pt}
\centering
\includegraphics[width=0.9\textwidth]{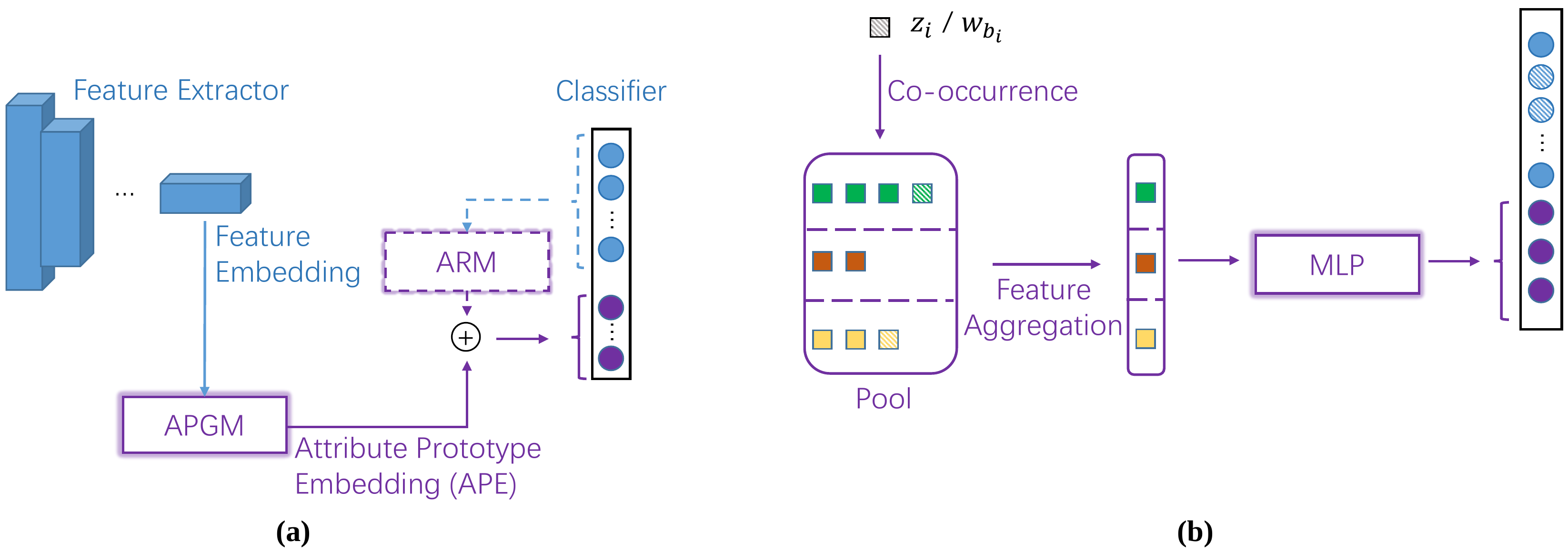}
\caption{ Overall Approach and Pipeline}
\label{architecture}
\end{figure*}

\subsection{Incremental Learning}
The problem we address here is also related to (but distinct from) incremental learning, also known as lifelong learning or continual
learning, where the model is given a sequence of tasks to learn. The main issue of incremental learning is to overcome catastrophic forgetting \cite{goodfellow2013empirical}, a phenomenon where the model performs poorly on the previous tasks after learning the new tasks. 



Two representative works of overcoming catastrophic forgetting are LwF \cite{li2018learning} and EWC \cite{kirkpatrick2017overcoming}. LwF aims to alleviate the catastrophic forgetting by exploiting knowledge distillation on the old output nodes of the model. EWC, on the other hand, evaluates each neurons importance for previous tasks using fisher information matrix and keep the important neurons close to their original values. The incremental few-shot problem we are facing here differs from the standard incremental setting where the learning of the novel classes is harder and catastrophic forgetting is less severe.

\section{Problem Setup}

Incremental few-shot attribute learning is essentially a \textbf{multi-label few-shot learning} problem. We assume that in phase one we are given a large-scale pedestrian attribute dataset $\mathcal{D}_{base} = 
\{(x_1,y_1),...,(x_{K_{base}},y_{K_{base}})\}, \quad y_i \in \{0,1\}^{N_{base}}$, with attributes $\mathcal{A}_{base}=\{a_1,...,a_{N_{base}}\}$, together with a model $\mathcal{M}_{base}$ which is well-trained on it. 

Then in phase two, we are provided with a new dataset $\mathcal{D}_{novel}$, in which novel attributes $\mathcal{A}_{novel}=\{a_1,...,a_{N_{novel}}\}$ appear. The $\mathcal{D}_{novel}$ contains $K_{novel}$ examples annotated with $N_{novel}$ novel attributes where $K_{novel}$ is usually small, and the entire learning process is termed as $N$-way $K$-shot learning. 

Our goal is to extend the $\mathcal{M}_{base}$ to $\mathcal{M}_{ext}$ in the second phase so that it is able to recognize both $\mathcal{A}_{base}$ and $\mathcal{A}_{novel}$. It is worth noting that $\mathcal{A}_{novel}$ can be either absent or not annotated in $\mathcal{D}_{base}$. Unlike the normal multiclass incremental few-shot learning setting which assumes $\mathcal{D}_{base} \cap \mathcal{D}_{novel} = \phi$ , our setting only requires $\mathcal{A}_{base} \cap \mathcal{A}_{novel} = \phi$ and $\mathcal{A}_{novel}$ will not appear in $\mathcal{D}_{base}$, in which case the same example may appear both in $\mathcal{D}_{base}$ and $\mathcal{D}_{novel}$, but are annotated with $\mathcal{A}_{base}$ and $\mathcal{A}_{novel}$ respectively. In other words, it is more of \textbf{class incremental} rather than the \textbf{data incremental} setting.


\section{Proposed Method}
\subsection{Overview}

Our overall approach for incremental few-shot attribute learning is illustrated in Figure \ref{architecture}(a). We first start with a well-trained base model $\mathcal{M}_{base}$ (the blue part) with conceptually two parts: feature extractor $\mathcal{F}$ and classifier $\mathcal{W}$. 
We then introduce the APGM module which has no interference with the base model so that the performances on $\mathcal{A}_{base}$ will not be hampered. The APGM takes the feature embedding that contains the encoded information from multiple attributes as inputs, and produces the separated APE embedding for each attribute. Then the classification weights for novel attributes are produced using the APEs. Moreover, when $\mathcal{D}_{novel}$ is also annotated with base attributes, we propose the ARM, which takes the base attributes' classification weights as inputs, and works in a similar way as APGM to produce the novel classification weights. In order to facilitate few-shot learning, we train these modules with episodic training 
on $\mathcal{D}_{base}$, which will be discussed in the next section.





\subsection{Episodic Training}
The essence of meta learning involves two levels of learning. During the low-level training, tasks $\{ \mathcal{T}_t \}$ (such as an $N$-way $K$-shot classification) are sampled from $\mathcal{D}_{base}$ as support examples $\mathcal{D}_{support}$, on which the learner is guided by the meta-learner to learn. Then during the high-level training, the learner is evaluated on $\mathcal{D}_{query}$ (which is also from $\mathcal{D}_{base}$ and held out from $\mathcal{D}_{support}$) with some meta-objective to find out how well the meta-learner is capable of guiding the learner to generalize to new tasks, and then the meta-learner is optimized using the meta-objective (which is cross-entropy loss in our case). This whole process is called an \emph{episode} \cite{vinyals2016matching}.
By repetitively episodic training, which is essentially mimicking the expected few-shot setting, the meta-learner will gradually acquire the ability to generalize on few-shot task $\mathcal{T'}$ from $\mathcal{D}_{novel}$ held-out from $\{ \mathcal{T}_t \}$.

Following the idea of episodic training, we first sample $N$-way $K$-shot tasks from $\mathcal{D}_{base}$ as $\mathcal{D}_{support}$, and treat the sampled attributes from $\mathcal{A}_{base}$ as \emph{fake novel attributes} $\mathcal{A}_{fake\_novel}$. We then produce the classification weights  $\hat{W}_{fake\_novel}$ for $\mathcal{A}_{fake\_novel}$ under the guidance of APGM (and ARM). Then we evaluate the classification performances of $\hat{W}_{fake\_novel}$ on $\mathcal{D}_{query}$, and let the gradient backpropagated to update the APGM and ARM. 

The whole episodic training procedure on $\mathcal{D}_{base}$ is shown in Algorithm \ref{algorithm}. 



\subsection{Attribute Prototype Generator Module}
The basic pipeline of APGM is shown in Figure \ref{architecture}(b). Note that different from the multiclass classification case, where each feature embedding only belongs to a single class, here each feature embedding contains fused information of multiple attributes, and the goal of APGM is to make efficient transformation to separate the fused information so that the generated APE is discriminative enough for classification.

The APGM mainly consists of two components: a pool and an MLP. The pool contains $N_{novel}$ slots, and each slot stores all the relevant feature embedding for one novel attribute. The MLP serves as the parameterized meta-learner, which is optimized during the high-level learning.
A well-trained APGM works as follows. Suppose that we have $N_{novel}$ novel attributes $\mathcal{A}_{novel}$, where in Figure \ref{architecture}(b) $N_{novel}=3$. First, the $i^{th}$ example $(x_i,y_i) \in \mathcal{D}_{novel}$ are fed into the feature extractor to obtain the feature embedding $z_i$. With its annotated novel attributes $\{a_{i1},...,a_{il_i}\} \subseteq \mathcal{A}_{novel}$, we push $z_i$ into the slots corresponding to ${a_{i1},...,a_{il_i}}$, where in the figure, $z_i$ is pushed into the first and the third slots. After forwarding the whole few-shot dataset $\mathcal{D}_{novel}$, we aggregate features over each slot, and feed the aggregated embedding for each novel attribute into the MLP to produce the APE. The aggregation operation can have multiple choices, and we choose summation in our experiment. The whole forward process of APGM module during meta training is shown in Algorithm \ref{APGM}.

\begin{algorithm}[tb]
\SetAlgoLined
\SetKwInOut{Input}{Input}
\SetKwInOut{Output}{Output}
\Input{ Base dataset $\mathcal{D}_{base}$, Base model $\mathcal{M}_{base}$, Base attributes set $\mathcal{A}_{base}$,
Number of sampled fake novel attributes $N_{fake\_novel}$, Number of support examples $K_{support}$, Number of episodes $E$.}
\Output{$\mathcal{M}_{base}$ with trained APGM,ARM modules}
\While{\emph{episode} $\leq$ E}{
$\mathcal{A}_{fake\_novel}$ = Sample($\mathcal{A}_{base}$, $N_{fake\_novel}$) \\
$\mathcal{D}_{fake\_novel}$,$\mathcal{D}_{fake\_base}$ = Split($\mathcal{D}_{base}$) \\
// where $\mathcal{D}_{fake\_novel}$ contains at least one \\ // attribute from $\mathcal{A}_{fake\_novel}$ \\
$\mathcal{D}_{support}$
= Sample($\mathcal{D}_{fake\_novel}$, $K_{support}$) \\ 
$\mathcal{D}_{query}$ = $\mathcal{D}_{base} \setminus \mathcal{D}_{support}$ 

\For{$(x_i,y_i)$ in $\mathcal{D}_{support}$}{
$z_i$ = FeatureExtract($x_i$) \\
$Z_{support}.append(z_i)$
}
$\hat{W}_{APE}$ = APGM$(Z_{support}, D_{support})$ \\
$\hat{W}_{ARM}$ = ARM$(W_{fake\_base}, D_{support})$ \\
$\hat{W}_{fake\_novel}$ = $\hat{W}_{APE} + \hat{W}_{ARM}$ \\
$\hat{W}$ = concat$(W_{fake\_base}, \hat{W}_{fake\_novel})$ \\
// low-level learning

\For{$(x_j,y_j)$ in $\mathcal{D}_{query}$}{
$score =$ FeedForward$(x_j, [\mathcal{F}, \hat{W}])$ \\
update AGPM and ARM using meta-objective $\mathcal{L}(score, y_j)$ \\
// high-level update
}
}
\textbf{return} APGM, ARM module
\caption{$N$-way $K$-shot Episodic Training }
\label{algorithm}
\end{algorithm}

\begin{algorithm}[tb]
\SetAlgoLined
\SetKwInOut{Input}{Input}
\SetKwInOut{Output}{Output}
\Input{ Support set $D_{support} = \{(x_i, y_i)\}$,
 support features $Z_{support} = \{z_i\}$, fake novel attributes $\mathcal{A}_{fake\_novel}$.}
\Output{Attribute Prototype Embedding $\hat{W}_{APE}$ }
\For{ $z_i$ in $Z_{support}$}{
    Get $x_i$'s attributes $[a_{i1}, a_{i2}, ..., a_{il_i}]$ \\
    \For{$a_{ij}$ in $[a_{i1}, a_{i2}, ..., a_{il_i}]$}{
    slot[$a_{ij}$]$.append(z_i)$}
    
}
\For{$a_k$ in $\mathcal{A}_{fake\_novel}$}{
    $\hat{W}_{APE}[a_k]$ = MLP(Aggregate(slot[$a_k$]))
}

\textbf{return} $\hat{W}_{APE}$
\caption{APGM Module}
\label{APGM}
\end{algorithm}

\subsection{Attribute Relationship Module}
During the generation of attribute prototype generation, only the novel attributes $\mathcal{A}_{novel}$ in $\mathcal{D}_{novel}$ is needed. If we are further provided with the base attributes $\mathcal{A}_{base}$ on $\mathcal{D}_{novel}$, then we can exploit the well-trained base classification weights (the output node in blue) and attributes relationships for further improvements.

The ARM module shares a similar structure with APGM module with a disjoint set of parameters where the input to the ARM is the base classification weights $W_{base}$ (the output node in blue). When forwarding an example $(x_i,y_i)$, whose attributes are denoted as $\{a_{i1},...,a_{il}\}$, for each co-occurred base/novel attribute pair $(a_{ij},a_{ik}), a_{ij} \in \mathcal{A}_{base}$ and $a_{ik} \in \mathcal{A}_{novel}$, we push $w_{a_{ij}} \in W_{base}$ into $a_{ik}$'s corresponding slots. Then all the weights in the slots are aggregated and feed into the MLP. This process is in essence a weighted attention over $W_{base}$ determined by how often two attributes will co-appear on the same instance.

Finally, the output classification weights for novel attributes are the addition of APGM and ARM's outputs. During practical implementation, we further normalize the classification weights to be unit vector following \cite{qi2018low}, in which case the softmax classifier is equivalent to nearest neighbour with cosine distance, which will help stabilizing the meta-training.

\subsection{Inference}
During the second phase when the actual novel attributes come, we fix the meta-learner, and feed forward the whole $\mathcal{D}_{novel}$ through APGM (and ARM if possible) to produce the classification weights $W_{novel}$ for $\mathcal{A}_{novel}$. Then a complete model $\mathcal{M}_{ext}$ is obtained by concatenating $W_{base}$ and $W_{novel}$, where the feature extractor $\mathcal{F}$ remains unchanged. It is worth noting that unlike re-training or finetuning which entails several steps of backpropagation and optimization, our method \textbf{only requires forwarding $\mathcal{D}_{novel}$ once to generate the $\mathcal{M}_{ext}$}, which could be valuable for applications where models need to update on-device.

\subsection{Discussions: Relationships with Other Meta Learning Methods}
In this section, we discuss the relationships among our method and two state-of-the-art meta learning based methods \cite{snell2017prototypical,gidaris2018dynamic}. First we analyze by decoupling the whole network $\mathcal{M}$ into feature extractor $\mathcal{F}$ and classifier $\mathcal{W}$. In \cite{snell2017prototypical} the Prototypical Network adopts the entire feature embedding encoder $\mathcal{F}$ as meta-learner, where the classifier $\mathcal{W}$ is the non-parametric KNN, which indicates that the prototype is the feature embedding itself. Both \cite{gidaris2018dynamic} and our method use an extra parameterized weight generator as meta-learner, which outputs the classifier weights $\mathcal{W}$. However, the meta-learner in \cite{gidaris2018dynamic} is a Hadamard production with learnable parameters that transforms the feature embedding into $\mathcal{W}$, where our meta-learner is an Multi-Layer Perceptron (MLP). 

As we will illustrate in the experimental section, that simply modifying \cite{snell2017prototypical,gidaris2018dynamic} to their multi-label versions perform poorly in our setting.
For Prototypical Network, we argue that choosing $\mathcal{F}$ as the meta-learner in multi-label setting will not be able to learn class prototypes that generalize and transfer as they do in the multiclass case. This is because the same prototype can be attributed to multiple labels, thus it is almost impossible to represent prototype information from multiple labels within a single class prototype. Moreover, updating the $\mathcal{F}$ during episodic training will cause catastrophic forgetting on the base attributes.
By comparison, our method and \cite{gidaris2018dynamic} avoid this issue by \textbf{decoupling the meta-learner and the feature extractor $\mathcal{F}$}, so that $\mathcal{F}$ can remain unchanged during the second phase, thus keeping the performances on the base attributes. However, \cite{gidaris2018dynamic} uses Hadamard production with learnable parameters as meta network, which indicates that \textbf{the same set of coefficients is used for all attribute prototypes generation} and we argue that it is not discriminative enough to disentangle the fused information and produce the classification weights. Our meta network, however, is an MLP that provides different sets of coefficients for each class prototype generation, thus leading to more discriminative transformation and superior performances. 


\section{Experimental Results}
\subsection{Data Preparation}
We evaluate on two pedestrian attribute benchmark datasets PETA and RAP. We first modify these two datasets into incremental few-shot setting. To be specific, for $N$-way $K$-shot incremental few-shot learning, we first sample $N$ attribute groups (see \cite{zhao2018grouping}), then randomly choose one attribute from each of the selected $N$ groups to form $\mathcal{A}_{novel}$, and treat the remaining attributes as $\mathcal{A}_{base}$. 
We follow \cite{li2015multi} and \cite{li2016richly}, and divide the PETA and RAP into train/val/test set with 5 random partitions. All the reported results are the average results of these 5 partitions.
We then further divide each the train/val/test set into two parts: base and novel. An example is included in the novel sets if it contains at least one novel attribute, and is included in the base sets following the same rule.
We then randomly keep $K$ examples in the train$_{novel}$ set and discard the rest of the examples. During phase one, the model is trained on train$_{base}$/val$_{base}$, and test with test$_{base}$. During the second phase, the model is provided with the $K$-shot train$_{novel}$/val$_{novel}$, and test on the whole test set containing both base and novel attributes.

\subsection{Baselines}
\paragraph{Finetune with Different Learning Rate} is the most straight-forward method in dealing with few-shot settings. We fix the learning rate of classifier $\mathcal{W}$ to 0.001, and without specification, the \emph{lr} refers to the learning rate of the feature extractor $\mathcal{F}$.
\paragraph{Prototypical Network}\cite{snell2017prototypical} is a simple yet effective meta learning method. During the experiment, we change the metric from Euclidean distance to cosine similarity to cater for the multi-label setting.
\paragraph{DFWF}\cite{gidaris2018dynamic} is a recent state-of-the-art meta learning method. During the experiment, we also modify the original DFWF for the multi-label setting.

\subsection{Evaluation Metric}
In the experiment, we mainly focus on four metrics: (1) Class-centric mA: the mean average of each attribute's accuracy of positive and negative example. Specifically, we mainly focus on \textbf{mA$_{base}$} and \textbf{mA$_{novel}$} as they indicate how well the model can adapt to the novel attributes as well as remembering the old attributes in our class-incremental setting. (2) Instance-centric: the Precision, Recall and F1 calculated on each instance.
\begin{figure}[!h]
\centering
\includegraphics[width=0.48\textwidth]{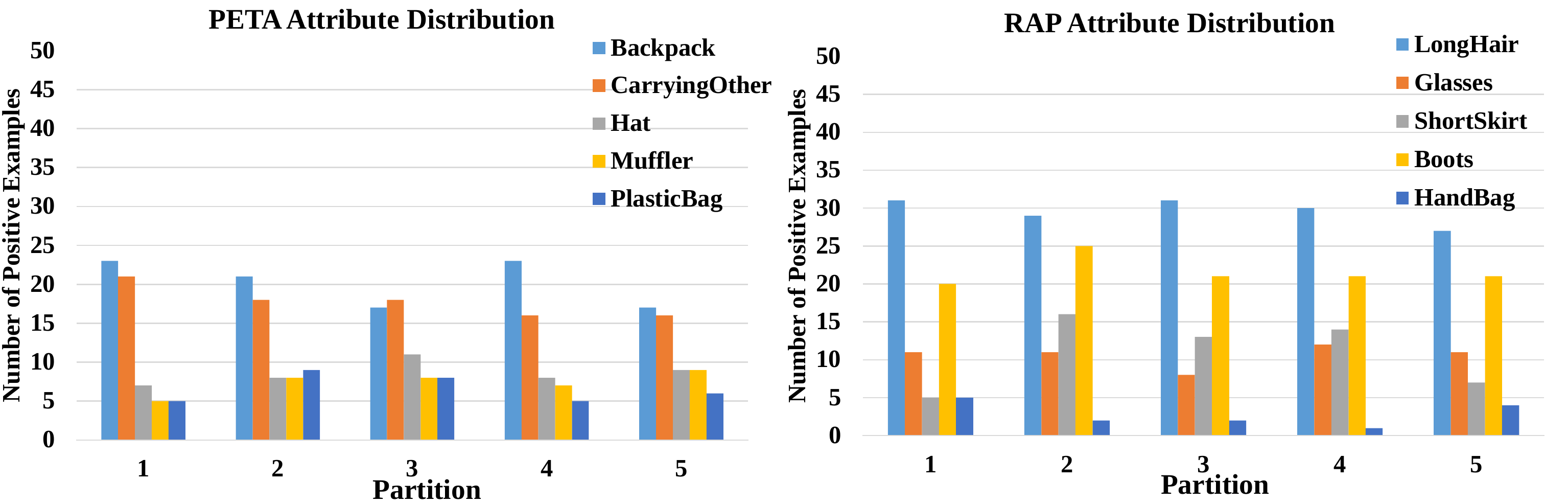}
\caption{Selected Novel Attributes Distribution of 5-way 50-shot Learning on PETA and RAP}
\label{distribution}
\end{figure}

\begin{table*}

  \centering
  \begin{tabular}{p{2.35cm}|p{1cm}<{\centering} p{1cm}<{\centering}p{1cm}<{\centering}p{1cm}<{\centering}p{0.8cm}<{\centering}|p{1cm}<{\centering}p{1cm}<{\centering}p{1cm}<{\centering}p{1cm}<{\centering}p{0.8cm}<{\centering}}
    \hline
    
    \multirow{2}{*}{\diagbox[width=2.8cm]{Method}{Metric}{Dataset}} & \multicolumn{5}{c|}{PETA } & \multicolumn{5}{c}{RAP } \\ [1ex] 
    \cline{2-11}
      &  mA$_{Base}$ & mA$_{Novel}$ & Precision & Recall & F1 & mA$_{Base}$ & mA$_{Novel}$ & Precision & Recall & F1 \\ [0.5ex]
    \hline
    Ft (lr=0 ) & $\downarrow$ 0.15 & 65.66 & 83.22 & 82.22 & 82.71 & $\downarrow$ 0.47 & 67.57 & 77.18 & 72.86 & 74.95 \\ [0.5ex] 
    Ft (lr=0.01) & $\downarrow$ 2.36 & 61.84 & 81.12 & 81.06 & 81.08 & $\downarrow$ 1.19 & 62.79 & 76.34 & 73.58 & 74.92 \\ [0.5ex]
    Ft (lr=0.001) & $\downarrow$ 0.24 & 64.12 & 83.10 & 82.05 & 82.57 & $\downarrow$ 0.39 & 66.38 & 77.19 & 72.97 & 75.02 \\ [0.5ex]
    Ft (lr=0.0001) & $\downarrow$ 0.12 & 65.69 & \textbf{83.24} & 82.11 & 82.67 & $\downarrow$ 0.44 & 67.47 & \textbf{77.24} & 72.98 & \textbf{75.05} \\ [0.5ex]
    Ft (lr=0.00001) & $\downarrow$ 0.13 & 65.52 & 83.23 & 82.28 & \textbf{82.75} & $\downarrow$ 0.46 & 67.67 & 77.22 & 72.93 & 75.01 \\ [0.5ex] 
    \hline 
    
    ProtoNet & 49.63 & 50.00 & 27.55 & 75.37 & 40.26 & 49.66 &  50.00 & 18.13 & 91.65  & 30.27 \\ [0.5ex] 
    DFWF  & \textbf{$\downarrow$ 0} & 54.46 &  70.42 & 82.51 &  75.78 & \textbf{$\downarrow$ 0} &  57.80 & 70.96 & 77.14 & 65.82  \\ [0.5ex] 
    \hline
    APGM (Ours) & \textbf{$\downarrow$ 0.0 } & \textbf{68.13} & 74.05 & \textbf{84.53} & 78.93 & \textbf{$\downarrow$ 0.0 } & \textbf{73.06} & 71.31& \textbf{76.95} & 74.02 \\ [0.5ex]    
    \hline
   
    \hline
  \end{tabular}
\caption{5-way 50-shot learning results on PETA and RAP benchmark. Ft (lr) refers to finetuning, and lr indicates the learning rate of $\mathcal{F}$. $\downarrow$ means decrease in mA$_{base}$ as compared to the base model.}
\label{table1}
\end{table*}

\subsection{Implemetation Details}
\paragraph{Base Network.}
Since we focus on the learning algorithm which is agnostic to the specific model, we adopt DeepMAR\cite{li2015multi} with ResNet50 as backbone as our base model $\mathcal{M}_{base}$ for its competitive performances and simplicity. 
\paragraph{Hyperparameters.}
In order to train our meta network, we adopt SGD and a small learning rate of $10^{-5}$,  with a momentum of 0.9 and weight decay of 0.0005. We choose the batch size to be 32, train for 800 episodes and choose the sampled $N_{fake\_novel}$ to be twice the expected $N_{novel}$.
During the meta training, we re-sample the support set if any $a_{fake\_novel}$ is absent in the $D_{support}$.

\paragraph{Alleviating Overfitting.}
In order to alleviate overfitting to a certain sampled task, we apply early stopping when the loss stops to decrease within each episode. During inference, we choose the model with the smallest training loss at the end of each episode as our final extended model.


\subsection{Results on 5-way 50-shot Learning}
We first experiment on PETA and RAP with 5-way 50-shot learning with two sets of randomly selected novel attributes, and the result is shown in Table \ref{table1}. Here only novel attributes are annotated on $\mathcal{D}_{novel}$. It is also worth to mention that although the total number of examples is 50, these attributes are highly imbalanced, where some attributes only contain less than 5 positive examples. The attribute distribution of the first five selected novel attributes is shown in Figure \ref{distribution}. Without specification, the rest of the results are reported on these five novel attributes.
The results from Table \ref{table1} show that compared to finetuning (Ft) with different learning rates, our method outperforms significantly in mA$_{novel}$ on both datasets, and holds comparable F1 on RAP while suffers a decrease in F1 on PETA. The drop in F1 mainly comes from the drop in Precision, and further regularization might be considered in future work. Besides, finetuning comes at a cost of forgetting on the base attributes where our method is able to retain the base model's performances. From the class incremental perspective, the class-centric metric mA$_{base}$/mA$_{novel}$ demonstrates our method's superiority in learning novel attributes on such few-shot learning setting. Moreover, finetuning involves several backpropagation and updating while our method requires much less time consumption. Finally, the results also accord with the discussions in Section 4.6, that the current meta learning method is not applicable in our setting where the ProtoNet acts almost like random guess, and DFWF also yields poor performances.

\begin{figure}[!h]
\centering
\includegraphics[width=0.49\textwidth]{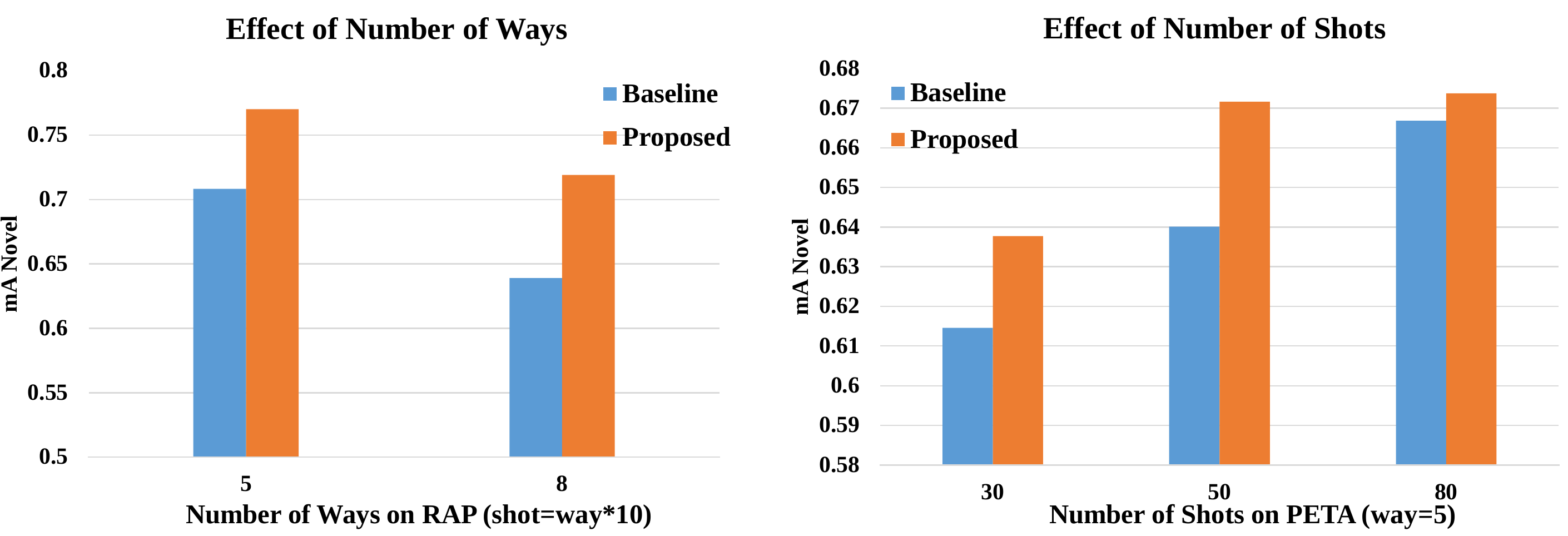}
\caption{Effect of Number of Shots and Ways}
\label{ablations}
\end{figure}

\subsection{Effect of ARM Module}

If we have access to the \emph{base attributes} of the \emph{novel data}, then we can apply ARM for further improvements. We experiment on PETA with 5-way 50-shot, and RAP with 8-way 80-shot where ARM can bring improvements in mA$_{novel}$ from 67.16\%/71.90\% to 68.05\%/72.88\%.

\subsection{Effect of Number of Ways and Shots}
We compare our method against finetuning (lr=0.0001) with different number of shots and ways. The results in Figure \ref{ablations} show that our method is able to consistently outperform the baseline on mA$_{novel}$. With different number of ways, our method is still able to generalize well to the novel attributes. However, the gap gets smaller as the number of shots increases, which implies that finetuning becomes a stronger baseline as the number of examples increases, and our method's superiority lies mostly on the few-shot scenarios.


During the experiments, we also empirically find that sampling more fake novel attributes than the expected number of ways, will yield a mA$_{novel}$ gain of $\sim1\%$. This training strategy is firstly adopted in \cite{snell2017prototypical}, which can be explained intuitively that 
making the sampled task more difficult (higher ways) during meta training is beneficial to the future generalization in the actual test case.




\section{Conclusion and Future Work}
In this paper, we raise the practical issue of incremental few-shot learning for pedestrian attribute recognition, a commonly existing problem which has gained little attention by the academic community. We tackle this challenging task by proposing a meta learning based framework, which only requires annotations on the novel attributes and one feed forward time to produce the adapted model. We evaluate our method against finetuning and other state-of-the-art meta learning methods, and argue that our method is able to rapidly adapt to the novel few-shot data with low resource consumption and yield competitive performances. We believe that this task is far from fully solved, and future work may involve designing meta-learner with stronger representation capability such as graph neural networks.

\section*{Acknowledgements}

The work is supported by National Key R\&D Program of China (2018YFC0806900), National Natural Science
Foundation of China (61571269) and National Basic Research Program
of China (2015CB352300).

\bibliographystyle{named}
\bibliography{ijcai19}
\end{document}